\documentclass[runningheads]{llncs}
\usepackage[T1]{fontenc}
\usepackage{graphicx}
\usepackage{booktabs}
\usepackage[misc]{ifsym}

\usepackage{multirow}
\usepackage{subcaption}
\usepackage{hyperref}
\usepackage{amsmath}
\usepackage{acro}
\DeclareAcronym{nas}{
  short=NAS,
  long=Neural Architecture Search,
}
\DeclareAcronym{hwnas}{
  short=HW-NAS,
  long=Hardware Aware NAS,
}
\DeclareAcronym{vit}{
  short=ViT,
  long=Vision Transformer,
}
\DeclareAcronym{cnn}{
  short=CNN,
  long=Convolutional Neural Network,
}
\DeclareAcronym{mcu}{
  short=MCUs,
  long=Micro-Controller Units,
}
\DeclareAcronym{mhsa}{
  short=MHSA,
  long=Multi-Head Self-Attention,
}
\DeclareAcronym{dnn}{
  short=DNN,
  long=Deep Neural Network,
}

\usepackage[dvipsnames,table,xcdraw]{xcolor}
\usepackage{tikz}
\usetikzlibrary{backgrounds}
\usetikzlibrary{calc,positioning,fit}
\usetikzlibrary{arrows.meta}
\usetikzlibrary{shapes.geometric}
\usetikzlibrary{bending}

\begin{document}

\title{Hybrid Convolution and Vision Transformer NAS Search Space for TinyML Image Classification}

\titlerunning{Hybrid CNN and ViT NAS Search Space for TinyML Image Classification}

\author{Mikhael Djajapermana\inst{1} 
\and
Moritz Reiber\inst{2}
\and
Daniel Mueller-Gritschneder\inst{3}
\and
Ulf Schlichtmann\inst{1}
}
\authorrunning{M. Djajapermana et al.}

\institute{Chair of Electronic Design Automation, Technical University of Munich, Munich, Germany
\email{\{mikhael.djajapermana,ulf.schlichtmann\}@tum.de}
\and
Department of Computer Science, University of T{\"u}bingen, T{\"u}bingen, Germany
\email{moritz.reiber@uni-tuebingen.de}
\and
Institute of Computer Engineering, TU Wien, Vienna, Austria
\email{daniel.mueller-gritschneder@tuwien.ac.at}
}

\maketitle

\begin{abstract}
Hybrids of Convolutional Neural Network (CNN) and Vision Transformer (ViT) have outperformed pure CNN or ViT architecture. However, since these architectures require large parameters and incur large computational costs, they are unsuitable for tinyML deployment. This paper introduces a new hybrid CNN-ViT search space for Neural Architecture Search (NAS) to find efficient hybrid architectures for image classification. The search space covers hybrid CNN and ViT blocks to learn local and global information, as well as the novel Pooling block of searchable pooling layers for efficient feature map reduction. Experimental results on the CIFAR10 dataset show that our proposed search space can produce hybrid CNN-ViT architectures with superior accuracy and inference speed to ResNet-based tinyML models under tight model size constraints.

\end{abstract}

\section{Introduction}

In recent years, \ac{vit} has become an increasingly popular \ac{dnn} architecture for computer vision tasks since it can outperform \ac{cnn}-based methods~\cite{dosovitskiy2020image,touvron2021training}. The main feature of \ac{vit} is the \ac{mhsa}, which allows efficient learning of global dependencies. However, \ac{vit}-based architectures typically require large training data since \ac{vit}s lack local inductive biases that are prevalent in \ac{cnn}s. Several works combined \ac{cnn} and \ac{vit} to form superior hybrid architectures \cite{mehta2021mobilevit,mehta2022separable}. \ac{nas} has been used to automate the hybrid \ac{cnn}-\ac{vit} design process~\cite{li2021bossnas,yang2023tinyformer}. A representative work is BossNAS~\cite{li2021bossnas}, which proposes an efficient \ac{nas} framework to search on a hybrid \ac{cnn} and Transformer search space with searchable downsampling positions. 

The performance growth of \ac{vit}s and hybrid \ac{cnn}-\ac{vit}s has been coupled with the increase in model size, computations, and memory requirements, thus hindering their deployment in resource-constrained edge devices or even \ac{mcu}, known as tinyML. The efficiency bottleneck in \ac{vit}s stems mainly from the \ac{mhsa}, whose computational and memory costs scale quadratically with the input resolution~\cite{vaswani2017attention}. Prior works reduced this quadratic complexity by introducing linear complexity \ac{mhsa}~\cite{wang2020linformer,liu2021swin,cai2022efficientvit}. A recent work ported \ac{vit}-based architecture to \ac{mcu}. For this, TinyFormer~\cite{yang2023tinyformer} proposes a framework consisting of SuperNAS, SparseNAS, and TinyEngine to generate and deploy efficient \ac{vit}-based architectures for image classification. 

In this paper, we present a hybrid \ac{cnn}-\ac{vit} search space for \ac{nas} that combines the capabilities of \ac{cnn} and \ac{vit} in tinyML model size scale for image classification. We consider diverse \ac{cnn} blocks based on popular architectures, such as ResNet~\cite{he2016deep} and MobileNetv2~\cite{sandler2018mobilenetv2}, and include the linear \ac{mhsa} in our \ac{vit} blocks. The \ac{cnn} blocks effectively capture local information, while the \ac{vit} blocks excel at learning long-range dependencies. Convolutions are often accompanied by pooling layers to extract the dominant features and reduce the feature map resolution. However, most \ac{nas} search spaces ignore the search for the parameters and type of the pooling layers. We argue that the pooling layers have a non-negligible impact on the model performance and introduce the searchable Pooling block in our search space. A good choice of pooling layers can improve the accuracy and efficiency of our hybrid \ac{cnn}-\ac{vit}s. We show that a \ac{nas} framework can easily leverage our proposed search space to produce a range of efficient hybrid architectures for tinyML image classification tasks. The generated models from our hybrid search space can be easily deployed with the help of existing \ac{dnn} compiler tool, such as TVM~\cite{chen2018tvm}.

In summary, the main contributions of this paper are as follows:
\begin{itemize}
    \item A more diverse search space for hybrid \ac{cnn}-\ac{vit}, containing diverse \ac{cnn} blocks and \ac{vit} blocks with ReLU-based linear \ac{mhsa} for better inference latency on \ac{mcu}. The models generated from our hybrid search space can be straightforwardly deployed using existing frameworks such as TVM.
    \item The novel Pooling block enables the creation of hybrid \ac{cnn}-\ac{vit} architecture with much-improved inference latency compared to other search spaces that do not consider searchable pooling layers.
\end{itemize}
The search space is implemented in the HANNAH NAS framework~\cite{gerum2022hardware}, using TVM to deploy the generated model to a RISC-V processor.
Experimental results on the CIFAR10 dataset~\cite{krizhevsky2009learning} show that the proposed hybrid search space allows the discovery of hybrid \ac{cnn}-\ac{vit}s that can outperform ResNet-based tinyML models in terms of latency and accuracy under tight model size constraints.
\section{Related Works}
\label{sec:related_works}

\subsection{Neural Architecture Search (NAS)}
\label{sec:related_nas}

\ac{nas} is a method to automate and speed up the neural network architecture design process. Standard \ac{nas} comprises three parts: search space, search algorithm, and performance estimator~\cite{elsken2019neural}. The search space defines the set of possible building blocks to form the network. The search algorithm explores the search space, generating architecture candidates and selecting the top candidates based on the ranking given by the performance estimator. Many different search algorithms have been employed, such as evolutionary search~\cite{real2019regularized}, reinforcement learning~\cite{zoph2016neural}, and differentiable supernet-based methods~\cite{liu2018darts}. For tinyML applications, the generated models should be able to operate on devices with limited memory, occasionally even with tight latency requirements. To consider the target device and requirements, \ac{hwnas} searches for well-performing architectures under hardware-related constraints, such as model size and latency. \ac{hwnas} methods have successfully identified efficient architectures with better trade-offs than manual designs, e.g., in image classification~ \cite{howard2019searching,cai2018proxylessnas,cai2019once}. 
Many \ac{hwnas} methods target the deployment on edge devices.
MCUNet~\cite{lin2020mcunet} optimizes the search space to fit the hardware target, trains a supernet with all possible sub-networks, and performs an evolutionary search to find the best sub-network for the target \ac{mcu}.
$\mu$NAS~\cite{liberis2021munas} uses Aging Evolution~\cite{real2019regularized} to search for the best architecture considering accuracy, RAM usage, model size, and latency.
A differentiable \ac{nas} is employed in MicroNets~\cite{banbury2021micronets} to find networks with low memory usage and operation counts. MicroNets achieve state-of-the-art results on three tinyML benchmarks: visual wake words, audio keyword spotting, and anomaly detection. 
HANNAH~\cite{gerum2022hardware} proposes a framework for automated hardware/software co-design of deep neural networks and neural network accelerators. HANNAH finds the best networks using an evolutionary search with randomized scalarization~\cite{paria2020flexible} based on power consumption and accuracy on audio classification tasks.
Once-for-all (OFA)~\cite{cai2019once} proposes the progressive shrinking technique to efficiently train the supernet. An evolutionary algorithm is performed to find a specialized sub-network from the OFA supernet based on the target hardware and latency constraints.

\subsection{Multi-Head Self-Attention (MHSA) in Transformer}
\label{sec:related_transformer}

Transformer~\cite{vaswani2017attention} is a popular deep-learning architecture for Natural Language Processing tasks, while Vision Transformer (ViT)~\cite{dosovitskiy2020image} is a variant designed for computer vision tasks. Their key feature is the \ac{mhsa} mechanism, which enables efficient learning of global dependencies. \ac{mhsa} operates on three different linear projections of the input $x$, namely $Q = x \cdot W_Q$, $K = x \cdot W_K$, $V = x \cdot W_V$. Then, the attention score is calculated as $A(x) = \text{softmax}\left(\frac{Q K^\top}{\sqrt{d}}\right) V$, where $d$ is the dimension of $K$. The attention score is multiplied by a linear weight $W_A$ to produce the output $\text{MHSA}(x) = A(x) \cdot W_A$.

A typical ViT encoder block is formed by stacking a \ac{mhsa} layer and a feed-forward layer with residual connection: $\text{ViT}_\text{enc}(x) = x + \text{FF}(x + \text{MHSA}(x))$. The feed-forward layer $\text{FF}$ consists of two linear weights with ReLU in between: $\text{FF}(x) = \text{ReLU}(x \cdot W_1) \cdot W_2$. The first linear weight $W_1$ expands the channel dimension, while the second linear weight $W_2$ reduces the dimension back to the input channel dimension.

\subsection{Convolution and Transformer NAS Search Space}
\label{sec:related_search_space}

BossNAS~\cite{li2021bossnas} proposes a \ac{nas} framework with the HyTra search space, which is a \ac{cnn} and Transformer search space with searchable downsampling positions. To create the architecture candidates, the HyTra search space allows \ac{cnn} and Transformer building blocks to be chosen in each layer. In contrast, we provide a macro skeleton to form the hybrid architecture by placing the \ac{cnn} blocks at the earlier layers and our hybrid \ac{vit} blocks at the subsequent layers. In BossNAS, the positional encoder performs the downsampling operation using convolutions. In our work, we also use convolutions as an implicit positional encoder, but we additionally introduce the Pooling block to explicitly downsample and reduce the feature map resolution. Furthermore, BossNAS does not target tinyML applications, while our work focuses on tinyML with much more restrictive model size and memory constraints. 

TinyFormer~\cite{yang2023tinyformer} proposes a framework to develop and deploy efficient Transformers on MCUs. The framework consists of three parts: SuperNAS, SparseNAS, and SparseEngine. TinyFormer employs a search space consisting of MobileNetV2 and Transformer blocks for image classification. In our search space, we consider not only the MobileNetV2 block but also other popular \ac{cnn} building blocks. For our Transformer block, we additionally consider ViT blocks with ReLU-based linear \ac{mhsa} inspired by \cite{cai2022efficientvit}. Our diversified search space enables the search for architectures with better trade-offs. TinyFormer performs well on MCUs but requires a specialized inference engine for deployment. In contrast, our model does not require a specialized inference engine for deployment while still providing inference speed benefits.

A well-designed search space may enable efficient search. For this, architectural properties from popular, well-performing models can be incorporated to reduce the search space size and simplify the search \cite{chen2021searching}. This work explores the design of hybrid \ac{cnn}-\ac{vit} search space for \ac{nas} to generate efficient hybrid architectures for tinyML image classification.
\section{Hybrid \ac{cnn}-\ac{vit} Search Space with Pooling Block}
\label{sec:proposed_method}

\begin{figure}[t]
\centering
\includegraphics[width=\textwidth]{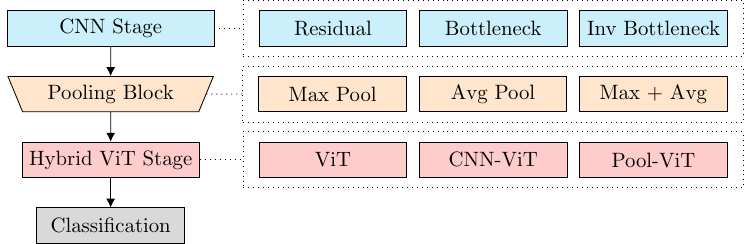}
\caption{Overview of our proposed hybrid search space.}
\label{fig:cnnvit_space}
\end{figure}

This section presents the proposed hybrid \ac{cnn}-\ac{vit} search space. As illustrated in \autoref{fig:cnnvit_space}, our hybrid search space provides a macro architecture skeleton that consists of a \ac{cnn} stage, a Pooling block, a Hybrid \ac{vit} stage, and a classification block.
The CNN stage operates as the local feature extractor, while the Hybrid \ac{vit} stage learns to extract global features. Similar to CPVT~\cite{chu2021conditional} and BossNAS~\cite{li2021bossnas}, our \ac{cnn} stage also serves as the positional encoder for the subsequent Hybrid \ac{vit} stage.
Our search space has a hierarchical structure where each \textit{stage} consists of several \textit{blocks} of its type, and the optimal number of blocks needs to be searched. The CNN and the Hybrid \ac{vit} blocks are described further in \autoref{sec:cnn_blocks} and \autoref{sec:vit_blocks}, respectively. The Pooling block (\autoref{sec:pooling_blocks}) consists of searchable pooling layers to reduce the feature map resolution and, thus, the computation costs of the subsequent layers. Finally, the classification block outputs the predicted probabilities of the classification classes. The classification block consists of an average pooling layer and a  linear layer without searchable architectural parameters.

Our search space covers many hybrid \ac{cnn}-\ac{vit} designs with diverse structures and complexities. In particular, the blocks within a stage are not forced to be uniform, i.e., one block in a stage may choose different parameters from another. Furthermore, each \ac{cnn}, Pooling, and Hybrid \ac{vit} block contains several independent searchable parameters.

\subsection{\ac{cnn} Block}
\label{sec:cnn_blocks}

\begin{figure*}[t]
\centering
\begin{subfigure}[b]{.33\textwidth}
  \centering
  \includegraphics[width=0.825\textwidth]{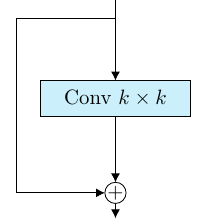}
  \caption{Residual.}
  \label{fig:residual}
\end{subfigure}%
\begin{subfigure}[b]{.33\textwidth}
  \centering
  \includegraphics[width=0.825\textwidth]{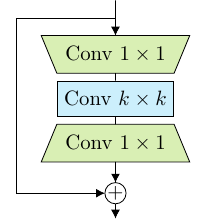}
  \caption{Bottleneck.}
  \label{fig:bottleneck}
\end{subfigure}%
\begin{subfigure}[b]{.33\textwidth}
  \centering
  \includegraphics[width=0.825\textwidth]{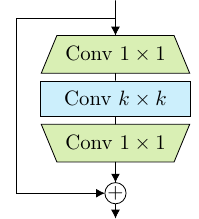}
  \caption{Inverted bottleneck.}
  \label{fig:inv_bottleneck}
\end{subfigure}
\caption{Types of CNN Blocks. The bottleneck structure is based on ResNet \cite{he2016deep}, while the inverted bottleneck is based on MobileNetV2 \cite{sandler2018mobilenetv2}.}
\label{fig:cnn_blocks}
\end{figure*}

We consider three types of \ac{cnn} blocks based on the structure found in popular \ac{cnn}-based models, such as ResNet~\cite{he2016deep} and MobileNetV2~\cite{sandler2018mobilenetv2}: 1) Residual block, 2) Bottleneck block, 3) Inverted bottleneck block. These block types are illustrated in \autoref{fig:cnn_blocks}.
Each \ac{cnn} block can search for its kernel size, stride, output channel, and group. When group $> 1$, the \ac{cnn} block performs the grouped convolution operation. The inverted bottleneck block is implemented with the depth-wise convolution~\cite{howard2017mobilenets}. For the residual connection, a $1 \times 1$ convolution operation is placed on the skip branch to ensure that the outputs of the two branches are of compatible shape for the addition.

\subsection{Pooling Block}
\label{sec:pooling_blocks}

\tikzset{%
    >={Latex[width=1.5mm,length=1.5mm]},
    invtrapez/.style={
        trapezium, trapezium angle=-67.5, draw=black,
        minimum width=1.55cm, minimum height=0.5cm,
        text centered
    },
    pool/.style = {invtrapez, fill=orange!20},
    circleplus/.style = {
        circle, fill=white, draw=black, minimum height=0.5cm, inner sep=0pt
    }
}

\begin{figure}[t]
\centering
\begin{subfigure}[b]{.33\textwidth}
  \centering
    \begin{tikzpicture}[remember picture,align=center,baseline=(current bounding box.center)]
    \node(max_pool) [pool]{Max Pool};
    \node(max_out) [circle, minimum size=0cm, inner sep=0pt, below=0.35cm of max_pool]{};
    \draw[->] (max_pool.south) -- (max_out);
    \end{tikzpicture}
  \caption{Max Pooling.}
  \label{fig:max_pool}
\end{subfigure}%
\begin{subfigure}[b]{.33\textwidth}
  \centering
    \begin{tikzpicture}[remember picture,align=center,baseline=(current bounding box.center)]
    \node(plus) [circleplus]{+};
    \node(combine_out) [circle, minimum size=0cm, inner sep=0pt, below=0.35cm of plus]{};
    \draw[->] (plus.south) -- (combine_out);
    \end{tikzpicture}
  \caption{Combination.}
  \label{fig:combination_pool}
\end{subfigure}%
\begin{subfigure}[b]{.33\textwidth}
  \centering
    \begin{tikzpicture}[remember picture,align=center,baseline=(current bounding box.center)]
    \node(avg_pool) [pool]{Avg Pool};
    \node(avg_out) [circle, minimum size=0cm, inner sep=0pt, below=0.35cm of avg_pool]{};
    \draw[->] (avg_pool.south) -- (avg_out);
    \end{tikzpicture}
  \caption{Average Pooling.}
  \label{fig:avg_pool}
\end{subfigure}%
\begin{tikzpicture}[remember picture,overlay]
    \draw[->] (max_pool.east) -- node [midway,above,align=center]{$\cdot 0.5$} (plus.west);
    \draw[->] (avg_pool.west) -- node [midway,above,align=center]{$\cdot 0.5$} (plus.east);
\end{tikzpicture}
\caption{Pooling Block with three types of pooling operators: (a) Max pooling, (c) Average pooling, and (b) a combination of Max and Average pooling.}
\label{fig:pooling_block}
\end{figure}
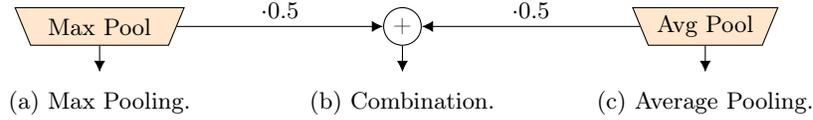

As illustrated in \autoref{fig:pooling_block}, we consider three pooling operations: 1) max pooling, 2) average pooling, and 3) a combination of max and average pooling by averaging their outputs. 
The purpose of the Pooling block is to explicitly reduce the feature map resolution by selecting the most critical information using searchable pooling layers. This contrasts the \ac{cnn} blocks, which implicitly reduce the feature map resolution if a convolution stride larger than $1$ is applied. 

\subsection{Hybrid \ac{vit} Block}
\label{sec:vit_blocks}

\begin{figure}[t]
\centering
\begin{subfigure}[b]{.45\textwidth}
  \centering
  \includegraphics[width=0.95\textwidth]{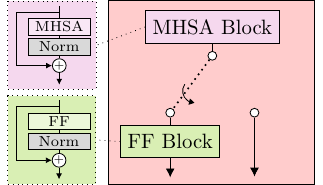}
  \caption{ViT Block.}
  \label{fig:vit_block}
\end{subfigure}%
\begin{subfigure}[b]{.25\textwidth}
  \centering
  \includegraphics[width=0.85\textwidth]{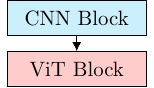}
  \caption{CNN-ViT Block.}
  \label{fig:cnn_vit_block}
\end{subfigure}%
\begin{subfigure}[b]{.25\textwidth}
  \centering
  \includegraphics[width=0.85\textwidth]{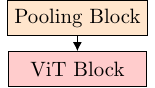}
  \caption{Pool-ViT Block.}
  \label{fig:pool_vit_block}
\end{subfigure}%
\caption{Types of Hybrid ViT Blocks. 
The (a) ViT Block comprises an MHSA Block and an optional Feed-Forward (FF) Block.}
\label{fig:hybrid_vit_block}
\end{figure}

The Hybrid \ac{vit} block always contains the \ac{vit} encoder block, with the optional \ac{cnn} or Pooling block preceding the encoder block.
Thus, we consider three types of Hybrid \ac{vit} Blocks: ViT, CNN-ViT, and Pool-ViT, as illustrated in \autoref{fig:hybrid_vit_block}. 
The \ac{vit} block (\autoref{fig:vit_block}) consists of a \ac{mhsa} sub-block and an optional feed-forward sub-block.     The \ac{mhsa} sub-block consists of a \ac{mhsa} layer and a normalization layer with a residual connection. The feed-forward sub-block follows the same design but uses a feed-forward layer instead of a \ac{mhsa} layer. 
The searchable parameters of the \ac{vit} block are the number of attention heads, the $Q$-$K$-$V$ dimensions, the choice to include the feed-forward block, and the feed-forward dimension. The feed-forward layer and the $Q$-$K$-$V$ projector in the \ac{mhsa} layer are implemented with a $1 \times 1$ convolution.  
Batch normalization~\cite{ioffe2015batch} is utilized instead of layer normalization since it can be folded into the preceding convolution in inference, offering a runtime advantage~\cite{cai2022efficientvit}.
For the activation function inside the feed-forward layer, we utilize ReLU~\cite{nair2010rectified} instead of Swish~\cite{ramachandran2017searching} or HardSwish~\cite{howard2019searching}, which are more commonly used in the recent \ac{vit} designs but may not be well-supported by the \ac{dnn} deployment tool. 

The \ac{vit} block is especially beneficial when placed after the Pooling block. We name this block combination the Pool-ViT block.
In the Pool-ViT block, the \ac{vit} block operates on a feature map with reduced resolution, which decreases the computational cost of the \ac{mhsa} operation that depends quadratically on its input resolution. 
We empirically show the benefits of the Pool-ViT blocks on inference efficiency and accuracy performance in \autoref{sec:experiments_results}.
This design is similar to the ResNet~\cite{he2016deep} design and light-weight \ac{vit} designs with pyramidal network structure~\cite{chu2021conditional,graham2021levit,pan2022edgevits}, which applies strong resolution reductions on its initial layers to reduce the computational costs of subsequent layers.

Moreover, inspired by EfficientViT~\cite{cai2022efficientvit}, we implement a ReLU-based linear \ac{mhsa} variant that avoids the Softmax function, as follows:
\begin{align*}
    &Q' = \text{ReLU}(x \cdot W_Q), \; K' = \text{ReLU}(x \cdot W_K), \; V = x \cdot W_V,
    \\
    &\text{MHSA}_{\text{linear}}(x) = Q' \cdot (K'^\top V) \cdot W_A 
\end{align*}
Avoiding Softmax improves the inference speed by removing the time-consuming calculation of the negative exponential~\cite{yang2023tinyformer}.
During the architecture search, the \ac{mhsa} sub-block inside our \ac{vit} block can choose either the standard \ac{mhsa} with Softmax or the ReLU-based linear \ac{mhsa} operation.
\section{Experiments}
\label{sec:experiments}

\subsection{Experiment Setup}
\label{sec:experiments_tools}

For evaluation, we implement our proposed hybrid \ac{cnn}-\ac{vit} search space in an existing \ac{nas} framework called HANNAH~\cite{gerum2022hardware}. HANNAH provides the tool to intuitively build complex search space designs and targets the tinyML use cases. We consider the image classification task with the CIFAR10 dataset \cite{krizhevsky2009learning}. We employ the evolutionary search in HANNAH to find a set of architecture candidates with the highest possible accuracy under a model size constraint of 100k parameters. During the search, the architecture candidates are trained for ${\sim}10$ epochs, and the validation accuracy is utilized as a performance proxy for the fully trained models. After the search is finished and the best candidate is found, we re-train the best candidate for 500 epochs to obtain its best accuracy performance. 

We compare our proposed hybrid search space defined in \autoref{sec:proposed_method} to other search space designs. As the baseline comparison, we build a ResNet-like search space by replacing the hybrid ViT stage with a CNN stage on our hybrid search space. Then, we define two separate modifications on our hybrid search space: 1) by removing the Pool-ViT block from the search space definition, and 2) by using only the ReLU-based linear \ac{mhsa} in the \ac{vit} block. In total, we consider four search space designs for our comparison and use the following notations for the search spaces: $\mathcal{R} =$ ResNet-like, $\mathcal{H} =$ proposed hybrid \ac{cnn}-\ac{vit}, $\mathcal{H}_{\text{WP}} = \mathcal{H}$ without the Pool-ViT block, $\mathcal{H}_{\text{L}} = \mathcal{H}$ with only linear \ac{mhsa}.

MLonMCU~\cite{van2023mlonmcu}, a TinyML deployment and benchmarking tool, is utilized to deploy the best model to an ETISS target simulating RISC-V processor~\cite{mueller2017extendable}. All generated models from HANNAH are in PyTorch. We convert the best-found models to ONNX format, which are then compiled using TVM and benchmarked with the help of MLonMCU. The benchmark results are presented in \autoref{sec:experiments_results}. 

\subsection{Evaluation and Comparison}
\label{sec:experiments_results}

\begin{figure}[t]
\centering
\begin{subfigure}[c]{0.55\textwidth}
  \centering
  \includegraphics[width=\textwidth]{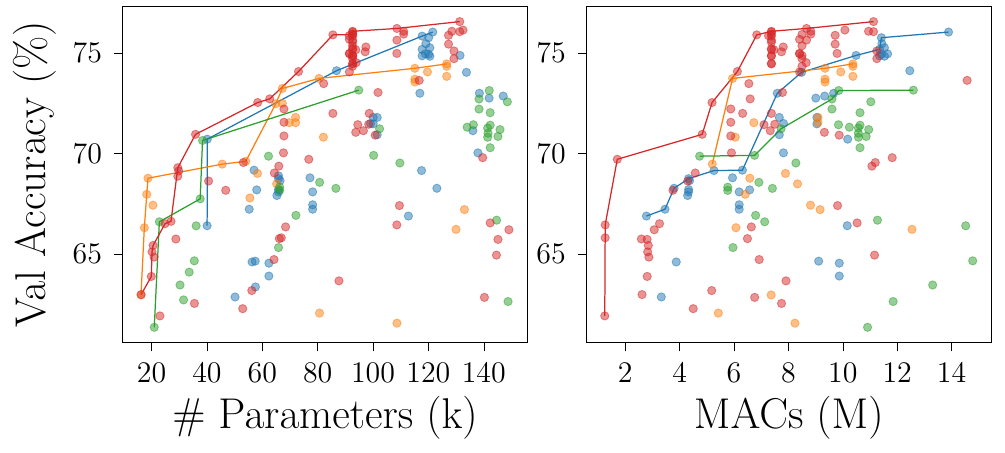}
\end{subfigure}%
\hfill
\begin{subfigure}[c]{0.45\textwidth}
  \centering
  \includegraphics[width=0.85\textwidth]{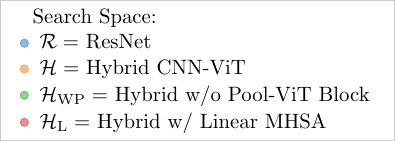}
\end{subfigure}%
\caption{Generated architecture candidates from four different search space designs on CIFAR10. The Pareto frontiers are connected with a line.}
\label{fig:candidate_search}
\end{figure}

\autoref{fig:candidate_search} shows the generated architecture candidates during the search with HANNAH using the ResNet-like search space, the proposed hybrid \ac{cnn}-\ac{vit} search space and its two modified versions. The hybrid \ac{cnn}-\ac{vit} models can perform competitively to the generated ResNet-like models. With enough parameters, the ResNet-like models can outperform the hybrid models. However, the hybrid architecture is more efficient in terms of model size and inference speed while maintaining competitive accuracy. In regions of $< 500$ kB model size (ROM), the hybrid architecture can achieve better inference speed and accuracy on CIFAR10 than the ResNet-based architecture. The complexity of the hybrid search spaces leads to a slightly longer search time than the ResNet-like search space.

\begin{table}[t]
\centering
\caption{Comparison of the best-found models from four search space designs defined in \autoref{sec:experiments_tools} on CIFAR10. The search cost is measured in GPU hours.}
\begin{tabular}{|l|cccc|}
\hline
Search Space & ~~~~$\mathcal{R}$~~~~ & ~~~~$\mathcal{H}$~~~~ & ~~$\mathcal{H}_{\text{WP}}$~~ & ~~~$\mathcal{H}_{\text{L}}$~~~
\\ \hline\hline
Accuracy
& 86.0\% & 87.1\% & 85.2\% & 87.3\% 
\\ \hline
\# Parameters  & 86.8k  & 80.5k  &  94.8k  &  85.5k 
\\ \hline
MACs         & 12.5M  & 5.9M   &  12.6M  &  6.8M  
\\ \hline
ROM          & 420 kB & 401 kB & 475 kB  & 422 kB 
\\ \hline
RAM          & 293 kB & 178 kB & 381 kB  & 159 kB 
\\ \hline
Latency      & 1.81 s & 1.35 s & 2.38 s  & 1.15 s 
\\ \hline
Search Cost  & 15 h  & 23 h  &  21 h  & 22 h
\\ \hline
\end{tabular}
\label{table:nas_results}
\end{table}

We choose the best models from the generated candidates based on the validation accuracy and the constraints on the number of parameters. \autoref{table:nas_results} shows the performance of the best-found models from the four different search spaces on CIFAR10. We highlight the performance differences between $\mathcal{H}$ and $\mathcal{H}_{\text{WP}}$. The feature map compression with the Pool-ViT block proves beneficial, as the results indicate that models from the $\mathcal{H}$ search space can achieve better accuracy with fewer parameters and much lower latency than ones from $\mathcal{H}_{\text{WP}}$. The best model on CIFAR10 can be obtained from the $\mathcal{H}_{\text{L}}$ search space, i.e., when the ViT block considers only the ReLU-based linear \ac{mhsa}. 

\begin{table}[t]
\centering
\caption{Comparison of our best-found hybrid CNN-ViT generated from our hybrid search space $\mathcal{H}$ to existing manually designed models on CIFAR10.}
\begin{tabular}{|l|c|c|c|c|}
\hline
Model & ~~~Ours~~~   & ~~ResNet8~~ & MobileNetV1 & MobileNetV2 \\ \hline\hline
Accuracy
& 87.1\%  & 86.5\%  & 79.2\%  & 89.9\%  
\\ \hline
\# Parameters 
& 80.5 k  & 75 k    & 233 k   & 350 k 
\\ \hline
ROM     
& 401 kB  & 225 kB  & 540 kB  & 4.77 MB    
\\ \hline
RAM     
& 178 kB  & 108 kB  & 478 kB  & 4.8 MB
\\ \hline
Latency 
& 1.35 s  & 2.54 s  & 1.87 s  & 72.9 s
\\ \hline
\end{tabular}
\label{table:cnnvit_comparison}
\end{table}

We compare our best-found hybrid \ac{cnn}-\ac{vit} generated from $\mathcal{H}$ to existing manually designed models in \autoref{table:cnnvit_comparison}. The ResNet8 and MobileNetV1 models are taken from the MLPerf Tiny Benchmark \cite{banbury2021mlperf}. The MobileNetV1 was originally designed for the Visual Wake Word task, but we trained it on CIFAR10 for the comparison. Compared to ResNet8, our hybrid \ac{cnn}-\ac{vit} achieves higher accuracy and lower latency with comparable RAM usage. The ROM usage is larger for our hybrid model, but ROM is usually a less limiting factor than the RAM usage for tinyML deployment. Our hybrid model outperforms MobileNetV1 on every considered metric. The MobileNetV2 achieves the highest accuracy in our comparison but requires a much larger ROM and RAM than our hybrid model.

\begin{figure}[t]
\centering
\begin{minipage}[c]{.45\textwidth}
  \centering
  \includegraphics[width=0.8\textwidth]{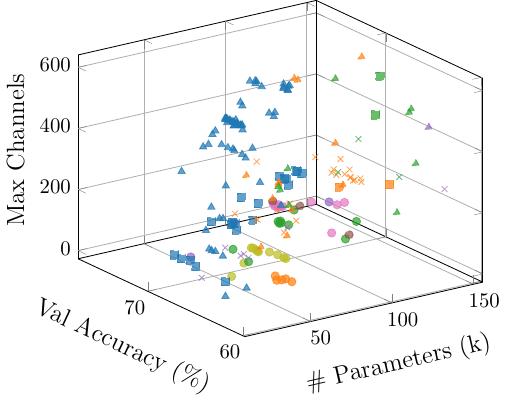}
  \label{fig:analysis_cifar10}
\end{minipage}%
\begin{minipage}[c]{0.25\textwidth}
  \centering
  \includegraphics[width=\textwidth]{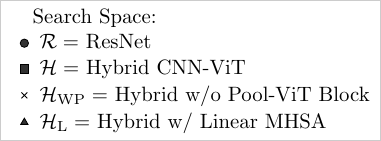}
\end{minipage}%
\begin{minipage}[c]{0.25\textwidth}
  \centering
  \includegraphics[width=0.5\textwidth]{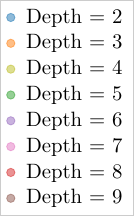}
\end{minipage}%
\caption{Relations between the architecture candidates' depth \& max channels (i.e., width) and their number of parameters \& accuracy.}
\label{fig:analysis_depth_width}
\end{figure}

Furthermore, we analyze the characteristics of the generated architecture candidates from the four different search spaces based on their depth and width, i.e., max channels. 
\autoref{fig:analysis_depth_width} shows that our hybrid models are generally shallower but wider than the generated ResNet-like models from $\mathcal{R}$. Furthermore, the candidates from the $\mathcal{H}_{\text{WP}}$ search space are mostly deeper but less wide than the ones from our proposed hybrid search space $\mathcal{H}$. Excessive use of pooling layers may cause the feature map size to be too small, resulting in information loss. This information loss is counterbalanced by increasing the number of channels, i.e., model width. The Pool-ViT block on our hybrid search space $\mathcal{H}$ implicitly inclines the \ac{nas} to create wide and shallow models. We leave a more thorough and formal analysis for future work.
\section{Conclusion}
\label{sec:conclusion}

This paper provides a search space for use in a \ac{nas} framework to design efficient hybrid \ac{cnn}-\ac{vit} network architectures with searchable Pooling blocks. Our search space design utilizes \ac{cnn} and \ac{vit} blocks to learn local and global information. The Pooling blocks effectively reduce the feature map resolution and, thus, the computation costs of the subsequent blocks. Implemented in the HANNAH \ac{nas} framework, our proposed search space can generate hybrid \ac{cnn}-\ac{vit} architectures that achieve superior accuracy and inference speed to the generated networks from the ResNet search space in tinyML model size regions. 

For future work, orthogonal techniques that compress existing networks, such as quantization and pruning, can be applied to further improve the efficiency of the generated networks. Knowledge distillation methods could improve the accuracy of the hybrid models under tight model size constraints. Furthermore, the model deployment process can be extended by considering NPU co-processing.

\begin{credits}
\subsubsection{\ackname} 
This work has been developed in the project MANNHEIM-FlexKI, funded by the German Federal Ministry of Education and Research (BMBF) under contract no.01IS22086L.

\subsubsection{\discintname}
The authors have no competing interests to declare that are relevant to the content of this article. 
\end{credits}

\bibliographystyle{splncs04}
\bibliography{references}

\end{document}